\documentclass[a4paper,fleqn]{cas-dc}

\usepackage[numbers]{natbib}
\usepackage{soul} % 文本修饰
\usepackage{url} % 插入网址
\usepackage{caption} % 图表标题格式调整

\usepackage{booktabs} % 表格线条控制
\usepackage{graphicx} % 插入图片
\usepackage{epsfig} % 插入EPS图片
\usepackage{float} % 浮动体控制
\usepackage{placeins} % 浮动体位置控制
\usepackage{xcolor} % 颜色支持
\usepackage{stfloats} % 插入浮动体
\usepackage{tabularx} % 表格控制
\usepackage{multirow} % 合并单元格
\usepackage{subcaption} % 子图表支持
\usepackage{pgfplots} % 绘制图表

\usepackage{inputenc}

\usepackage{amsmath}

\usepackage{algorithm}
\usepackage{algorithmic}

\usepackage{enumitem}

\usepackage{xstring}
\usepackage{xspace}
\usepackage{tikz} % 绘图支持
\usepackage{adjustbox} % 调整盒子大小
\pgfplotsset{compat=1.18}

\def\tsc#1{\csdef{#1}{\textsc{\lowercase{#1}}\xspace}}
\tsc{WGM}
\tsc{QE}
\tsc{EP}
\tsc{PMS}
\tsc{BEC}
\tsc{DE}
\begin{document}
\let\WriteBookmarks\relax
\def\floatpagepagefraction{1}
\def\textpagefraction{.001}

% % Short title
\shorttitle{MER}

% % Short author
\shortauthors{Zihan Zhou et~al.}

\title [mode = title]{Leveraging Motion Information for Better Self-Supervised Video Correspondence Learning}                      
% \tnotemark[1,2]

% \tnotetext[2]{}

\author[1]{Zihan Zhou}[style=chinese,
                        orcid=0009-0007-3650-5464]
\fnmark[1]
\cormark[1]
\ead{zhzhou@seu.edu.cn}

\credit{Conceptualization of this study, Methodology, Software}

\author[1]{Changrui Dai}[style=chinese]
\fnmark[1]
\ead{crdai@seu.edu.cn}
\credit{Writing – original draft, Investigation, Formal analysis}

\author[1]{Aibo Song}[style=chinese]
\ead{absong@seu.edu.cn}
\credit{Data curation, Writing - Original draft preparation}

\author[1]{Xiaolin Fang.}[style=chinese]
\cormark[2]
\ead{xiaolin@seu.edu.cn}
\credit{Writing – review \& editing, Supervision, Project administration}

\affiliation[1]{organization={School of Computer Science and Engineering, Southeast University},
    addressline={No. 2 Southeast University Road}, 
    city={Nanjing},
    % citysep={}, % Uncomment if no comma needed between city and postcode
    postcode={211189}, 
    state={Jiangsu},
    country={China}}

\cortext[cor1]{Corresponding author}
\cortext[cor2]{Principal corresponding author}

\fntext[fn1]{Equal contribution.}

\begin{abstract}
Self-supervised video correspondence learning depends on the ability to accurately associate pixels between video frames that correspond to the same visual object. However, achieving reliable pixel matching without supervision remains a major challenge. To address this issue, recent research has focused on feature learning techniques that aim to encode unique pixel representations for matching. Despite these advances, existing methods still struggle to achieve exact pixel correspondences and often suffer from false matches, limiting their effectiveness in self-supervised settings. 

To this end, we explore an efficient self-supervised Video Correspondence Learning framework (MER) that aims to accurately extract object details from unlabeled videos. First, we design a dedicated Motion Enhancement Engine that emphasizes capturing the dynamic motion of objects in videos. In addition, we introduce a flexible sampling strategy for inter-pixel correspondence information (Multi-Cluster Sampler) that enables the model to pay more attention to the pixel changes of important objects in motion. Through experiments, our algorithm outperforms the state-of-the-art competitors on video correspondence learning tasks such as video object segmentation and video object keypoint tracking.
\end{abstract}

% \includegraphics{figs/grabs.pdf}

\begin{highlights}
\item We propose a Motion Enhancement Engine to guide self-supervised video correspondence learning by emphasizing moving targets.
\item We propose a multi-cluster sampler that aims to improve pixel correspondence accuracy in complex video scenes without pixel label annotations through multi-round and multi-scale similarity sampling.
\item We introduce a fusion loss function to assign adaptive pixel weights, which improves the object tracking performance.
\item Our method effectively enhances self-supervise pixel correspondence learning, outperforming prior method sin video object segmentation and key point tracking.
\item The representations learned by the network through self-supervision methods show strong generalization ability in multiple video communication tasks, achieving comparable results to supervised methods.
\end{highlights}

% Keywords
\begin{keywords}
Self-Supervised Learning \sep 
Correspondence Learning \sep
Video Object Segmentation \sep
Pose Keypoint Tracking \sep
\end{keywords}

\maketitle

\section{Introduction}
Unlike supervised video Correspondence learning, self-supervised learning does not require intensive pixel-level annotation of video targets, which has significant advantages.
Leveraging the inherent temporal and spatial coherence of natural videos, self-supervised learning has emerged as a viable approach for establishing pixel correspondences between consecutive frames without the need for labeled data. This paradigm allows the model to effectively capture and learn distinctive pixel-level features of target objects within video frames, enhancing its ability to track correspondences over time.
Building on this foundation, related work in this area has garnered significant attention in domains such as surveillance\cite{satkin20133dnn}, autonomous driving\cite{lin2018architectural}, and behavior analysis\cite{Hu_2018_ECCV}. Furthermore, the abundance of natural video data enhances the model's generalization capabilities, potentially surpassing supervised methods in terms of data availability and applicability.

In the realm of existing \textit{self-supervised temporal correspondence learning} techniques, two predominant methods have been commonly employed to attain the desired objective. These methods include \emph{frame reconstruction}, where each pixel in a `query' frame is reconstructed by identifying and assembling relevant pixels from adjacent frame (s)  \cite{Lai_2020_CVPR,lai2019self,vondrick2018tracking}, and \emph{cycle-consistent tracking}, which encourages pixels or patches to return to the same location after one cycle of forward and backward tracking \cite{jabri2020space,li2019joint}.

It is important to emphasize that some previous studies on self-supervised video correspondence learning using pixel similarity have overlooked a key aspect of frame reconstruction. Specifically, they fail to take into account the different importance of pixel information in different regions of the video frame. Instead of acknowledging these differences, many methods treat all pixels uniformly, which we believe limits the model's ability to effectively capture the characteristics of target pixels through correspondences. This oversight limits the potential in accurately reconstructing frames and identifying meaningful pixel relationships in self-supervised scenarios.

We believe that each pixel in a video frame does not contribute equally to correspondence learning. For example, moving objects and "foreground" regions containing targets should be inherently more important than "background" regions. These regions deserve more attention.

To address the shortcomings present in previous work and to enable the model to genuinely focus on valuable parts, we propose a new \textbf{Motion Enhancement Engine} and \textbf{Multi-Cluster Sampler} framework. Furthermore, we have designed \textbf{Fused loss function} to enhance the model's performance. Specifically:

\begin{enumerate}
    \item \textbf{First}, we utilize pixel motion trend, which provides pixel offset information, to fuse with the pixel features of the original video frames. Additionally, we design a value extraction network to highlight moving targets within video sequences. This approach allows our self-supervised model to concentrate on valuable target information while minimizing interference from the background.
    \item \textbf{Second}, we propose a multi-round, multi-scale pixel similarity sampling algorithm that leverages clusters obtained through pixel clustering to indicate pixel correspondence matching. This approach enhances pixel-to-pixel accuracy in frame reconstruction and ensures robust performance during global similarity calculations for long-term tracking.
    \item \textbf{Last}, we refine the loss function by incorporating a weight map that assigns variable weights to pixels during loss calculation. This adjustment aligns the loss function more precisely with the object tracking task, where the quality of the target region is of utmost importance.
\end{enumerate}

\begin{figure}[h]
\centering
\begin{tikzpicture}
\begin{axis}[
    width=0.9\linewidth, % 设置图的宽度为0.9倍文字宽度
    xlabel={Number of pixel-level annotations (log scale)},
    ylabel={$\mathcal{J} \& \mathcal{F}$ (Mean)},
    xmode=log, % 使用对数坐标
    xmin=1, xmax=10000000, % 设置x轴范围    
    xticklabel style={align=center, text width=1.5cm, anchor=south, yshift=-3.0ex},
    xticklabels={$0$, $10^1$, $10^2$, $10^3$, $10^4$, $10^5$, $10^6$, $10^7$},
    ymin=63, ymax=87, % 设置y轴范围
    grid=both,
    grid style={dotted},
    enlargelimits={abs=0.1},
    only marks, % 仅绘制散点
    mark size=1.8, % 散点大小
    axis line style={draw=none}    
]
        \addplot[color=blue, mark=*] table {
        x   y
        1   65.5    
        1   66.7    
        1   68.3       
        1   70.3    
        1   71.4    
        1   72.1
        1   74.5
        };
        \addplot[only marks, mark=*, mark size=1.6, fill=red, draw=red] table{
        x   y
        1   76.5   
        };
        \addplot[color=yellow,mark=*] table {
        x   y
        5000000 85.4    
        2000000 82.8    
        3000000 81.7
        };
        \addplot[color=green,mark=*] table {
        x   y
        300000 75.9  
        600000 74.6  
        200000 75.3
        };
        \addplot[color=purple,mark=*] table {
        x   y
        200000 70.0 
        30000 66.7   
        };
        \node[anchor=west, font=\tiny] at  (axis cs:1,   65.5) {MAST\cite{Lai_2020_CVPR}};
        \node[anchor=west, font=\tiny] at  (axis cs:1,   66.7) {VFS\cite{xu2021rethinking}};
        \node[anchor=west, font=\tiny] at  (axis cs:1,   68.3) {CRW\cite{jabri2020space}};
        \node[anchor=west, font=\tiny] at  (axis cs:1,   70.3) {CLTC\cite{jeon2021mining}};
        \node[anchor=west, font=\tiny] at  (axis cs:1,   71.4) {DINO\cite{caron2021emerging}};
        \node[anchor=west, font=\tiny] at  (axis cs:1,   72.1) {LIIR\cite{li2022locality}};
        \node[anchor=west, font=\tiny] at  (axis cs:1,   74.5) {MASK-VOS\cite{li2023unified}};
        \node[anchor=west, font=\tiny] at  (axis cs:1,   76.5) {Ours};
        \node[anchor=east, font=\tiny] at  (axis cs:5000000, 85.4) {STCN\cite{cheng2021rethinking}};
        \node[anchor=east, font=\tiny] at  (axis cs:2000000, 82.8) {EGMN\cite{lu2020video}};
        \node[anchor=east, font=\tiny] at  (axis cs:3000000, 81.7) {STM\cite{oh2019video}};
        \node[anchor=west, font=\tiny] at  (axis cs:300000, 75.9) {Fasttan\cite{huang2020fast}};
        \node[anchor=west, font=\tiny] at  (axis cs:600000, 74.6) 
        {AFB-URR\cite{liang2020video}};
        \node[anchor=east,font=\tiny] at  (axis cs:200000, 75.3) 
        {MHP-VOS\cite{xu2019mhp}};
        \node[anchor=west, font=\tiny] at  (axis cs:200000, 70.0) {AGAME\cite{johnander2019generative}};
        \node[anchor=west, font=\tiny] at  (axis cs:30000, 66.7) {RGMP\cite{oh2018fast}};
\end{axis}
\end{tikzpicture}
\caption{\textbf{Performance comparison} over DAVIS$_{17}$\protect\cite{perazzi2016benchmark}val. Our MER surpasses all existing self-supervised methods ($\mathcal{J} \& \mathcal{F}$ (Mean) : 76.5), and is on par with many fully-supervised ones trained with massive annotations.}
\label{niubi}
\end{figure}
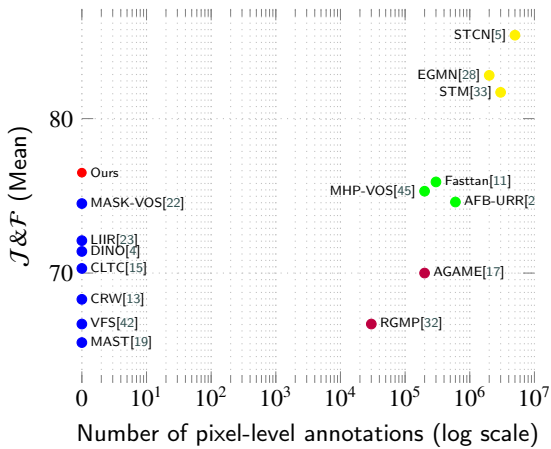

Overall, this framework of motion-enhanced frame reconstruction methods enables our self-supervised video correspondence learning model (MER) to achieve superior performance and outperform other methods in video object fine-grained tracking tasks such as video object segmentation and keypoint tracking. Notably, without the need for adaptation, the learned representations are shown to be effective for a variety of correspondence-related tasks, including video object segmentation and video object keypoint tracking. MER consistently surpasses state-of-the-art self-supervised techniques in this context and, in specific instances, exhibits performance on par with or adaptable to fully supervised methods, as exemplified in Fig.~\ref{niubi}.

\section{Related Work}
\subsection{Self-Supervised Temporal Correspondence Learning}
In the realm of video processing, correspondence learning is pivotal for various tasks such as video segmentation \cite{Hu_2018_ECCV}, flow estimation \cite{dosovitskiy2015flownet,zhao2022global}, and object tracking \cite{bertinetto2016fully}. Unlike supervised correspondence learning, self-supervised correspondence learning has leveraged rich, open, and unlabeled data sources \cite{2018Geometry,kim2019self,vondrick2018tracking} to learn the embedded correspondence relationships within videos, achieving significant progress.

Approaches to self-supervised temporal correspondence learning can be broadly classified into two categories. The first class of methods \cite{Lai_2020_CVPR,lai2019self,vondrick2018tracking} formulate a colorization proxy task, aiming to reconstruct a query frame from an adjacent frame, leveraging their temporal correspondence. The second class of methods \cite{jabri2020space,li2019joint,2019Unsupervised} involve forward and backward tracking, aiming to penalize inconsistencies between the initial and final positions of tracked pixels or regions. The central concept of cycle-consistency, as seen in these methods, is also adopted in self-supervised tracking \cite{2019Unsupervised,zheng2021learning}, optical flow estimation \cite{2018UnFlow,2018DF}, and depth estimation \cite{2018Unsupervised,2018GeoNet}.

Despite their great success, these methods face challenges in three key aspects necessary for robust correspondence matching: instance discrimination, location awareness, and error propagation. In this work, we propose to integrate additional object motion information to enhance instance recognition, encode location information into training, and reduce error propagation in pixel correspondence learning through flexible sampling strategies such as multi-scale sampling methods.

\subsection{Optical flow estimation}
The classic problem of optical flow estimation,  which involves predicting per-pixel motion between two frames,  has recently seen advancements in the use of synthetic graphics data for supervised training \cite{dosovitskiy2015flownet,mayer2016large}.  One of the pioneering deep learning methods to address end-to-end optical flow learning was FlowNet \cite{dosovitskiy2015flownet}.  This breakthrough research served as an inspiration for a plethora of other techniques,  including PWC-Net \cite{sun2018pwc},  LiteFlowNet3 \cite{hui2020liteflownet3}, and the more recently proposed and more effective RAFT \cite{teed2020raft}.  Many of these methods make use of cost volumes to determine pixel matches.  Among them, GMFlowNet\cite{zhao2022global} stands out by incorporating multi-scale correlation volumes and iterative flow refinements,  leading to superior performance.  

After many experiments, we used optical flow as a tool to carry the motion information of the target in the video and fused it with the video frames, introducing motion information into each static video frame. However, we believe that the introduction of optical flow may also bring other redundant information, which may affect the accuracy and stability of feature extraction. Therefore, we also designed the Value Extraction Network to ensure that the introduction of optical flow has a positive effect without affecting overall performance.

\subsection{Video Object Segmentation}
Video object segmentation is an important practice for video correspondence learning. Initial approaches to Video Object Segmentation (VOS) relied heavily on online finetuning during the first frame, a process known for its slow inference speeds and has since been phased out \cite{maninis2018video, caelles2017one}. To address the need for speed, researchers have introduced a variety of faster methodologies. These include improved online learning algorithms \cite{yang2018efficient}, inference using Markov Random Field (MRF) graphs \cite{bao2018cnn}, Temporal Convolutional Neural Networks (CNNs) \cite{xu2019spatiotemporal}, capsule routing mechanisms \cite{duarte2019capsulevos}, advancements in tracking techniques \cite{wang2019fast}, embedding learning \cite{yang2020collaborative}, and space-time matching strategies \cite{oh2019video, cheng2021rethinking}. Notably, embedding learning and space-time matching share a common goal: to learn a deep, consistent feature representation of an object across a video sequence. Typically, embedding learning techniques are more restrictive, utilizing local search windows and strict one-to-one matching paradigms.

Learning VOS in a self-supervised manner is attractive in the VOS domain because it eliminates the substantial annotation cost required for fully supervised algorithms. Due to the lack of mask annotation, existing self-supervised methods often adopt the correspondence learning method mentioned in Section 2.1, and many excellent algorithm models such as LIIR\cite{li2022locality} and Spat\cite{li2023spatial} were born. Worth mentioning is Mask-VOS\cite{li2023unified}, which guides segmentation by clustering fake masks and promotes correspondence between pixels to learn correct results. It has some similarities with the work of this paper.

\section{Methodology}
\begin{figure*}[tp]
    \centering
    \includegraphics[width=0.99\linewidth]{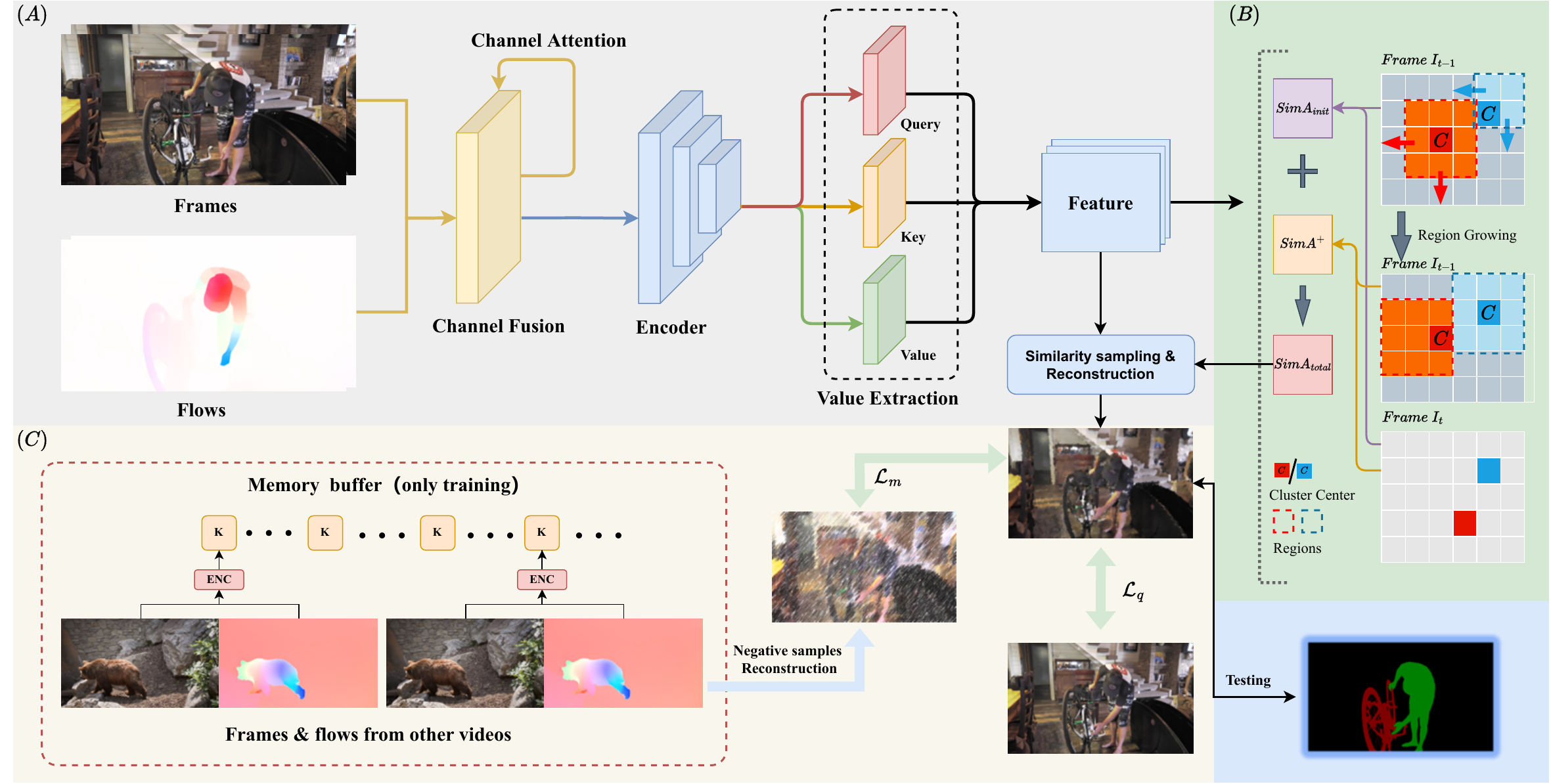}
    \caption{In the MER architecture, the current frame and its associated optical flow are jointly fed into the encoder as a self-concerned query frame.  This enables multi-scale similarity sampling and reconstruction using key values stored in memory.  During training, the original video frame is used as the self-supervised value.  Once the encoder is trained, we transition to using instance masks as values.}
    \label{fig:MER}
\end{figure*}

In order to improve the accuracy of pixel correspondence learning, we proposed a self-supervised model that aims to improve the accuracy of feature representation of moving targets by enhancing the ability to capture motion information (\textbf{Motion Enhancement Engine}), and designed a clustering-based multi-scale pixel similarity sampling strategy (\textbf{Multi-Cluster
Sampler}) from local to global. We call this model the motion-enhanced self-supervised pixel reconstruction model (the model architecture can be found in Figure ~\ref{fig:MER}), abbreviated as MER.

\subsection{Reconstruction of the Current Frame Using the Previous Frame}\label{base}
Let's review that in self-supervised Video Correspondence Learning, the goal is to develop robust object feature representations without labeled data. Videos are advantageous due to their inherent temporal information and spatio-temporal consistency. This allows us to use the continuity of pixel information, as pixels in one frame often correspond to pixels in the next. Based on this correlation, studies like \cite{lai2019self} have introduced reconstruction methods where each pixel in the current frame identifies its best match in the previous frame.

This reconstruction method incorporates attention mechanisms. Both the current frame and the previous frame undergo feature encoding and are projected into the pixel embedding space $\Phi:R^{H \times W \times 3} \rightarrow R^{h \times w \times c}$, yielding a triplet for each frame $I_{t}$, denoted as ${{Q_{t}, K_{t}, V_{t}}}$, where $Q_t = K_t = \Phi(I_t;\theta)$. During the training phase, the value $V_t$ corresponds to the original reference frame, while during inferring phase, it corresponds to the instance segmentation mask. 

\begin{figure}[ht]
    \centering
    \includegraphics[width=0.99\linewidth]{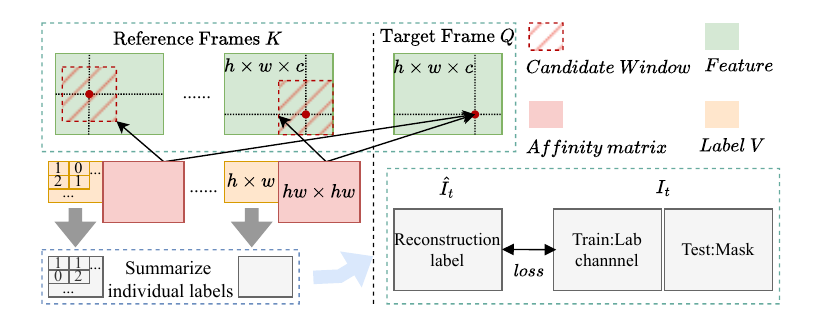}
    \caption{\textbf{Label reconstruction process.} For two frames with dimensions \( h \times w \), there will be a affinity matrix $ A \in \mathbb{R}^{hw \times hw}$ to represent the affinity between any two pixels. During the unsupervised training process, we use randomly selected $ \mathbf{Lab} $ channel as label. }
    \label{reconstruction fig}
\end{figure}

To reconstruct pixel \( i \) in the current frame \( I_t \), we perform similarity sampling from pixels in previous frames. Matching pixels contribute to reconstructing the pixel \( \hat{i} \) in the reconstructed frame \( \hat{I_t} \). This process involves using multiple previous frames and the current frame to compute affinity matrices and generate reconstruction results, summarized by:
\begin{equation}
    A^{ij}_{t, r} = \frac{\exp \langle Q^{i}_{t}, K^{j}_{r} \rangle}{\sum_{p} \exp \langle Q^{i}_{t}, K^{p}_{r} \rangle},
\label{base1}
\end{equation}
where \( A^{ij}_{t, r} \) denotes the affinity between the \( i \)-th feature of the target frame and the \( j \)-th feature of the \( r \)-th reference frame, with \( \langle Q^{i}_{t}, K^{j}_{r} \rangle \) representing their dot product (see Fig. \ref{reconstruction fig}, left).
\begin{equation}
    \hat{I}^{i}_{t, r} = \sum_{j} A^{ij}_{t, r} V^{j}_{r},
\label{base2}
\end{equation}
where \( \hat{I}^{i}_{t, r} \) is the reconstructed value at pixel \( i \) using features from the \( r \)-th reference frame, and \( V^{j}_{r} \) is the value associated with the \( j \)-th feature in the \( r \)-th reference frame.
All reconstruction results are then fused to reduce errors:
\begin{equation}
    \hat{I}_{t} = \textit{M}\left(\{\hat{I}^{i}_{t, r}\}_{r=1}^{R} \text{ for each pixel } i\right),
\label{fused i}
\end{equation}
where \( \hat{I}_{t} \) is the final reconstructed frame, and the function \( \textit{M} \) selects the most frequent value among all reconstructed results \( \hat{I}^{i}_{t, r} \) (see Fig. \ref{reconstruction fig}, right). Finally, the reconstruction loss is computed as:
\begin{equation}
    \mathcal{L}_{res} = ||I_t - \hat{I_t}||_2.
\end{equation}
In this context, previous frames are referred to as reference frames, and the current frame as the target frame, terminology that will be used throughout the paper.

\begin{figure*}[tp]
    \centering
    \begin{subfigure}[t]{\textwidth}
        \centering
        \includegraphics[width=\textwidth]{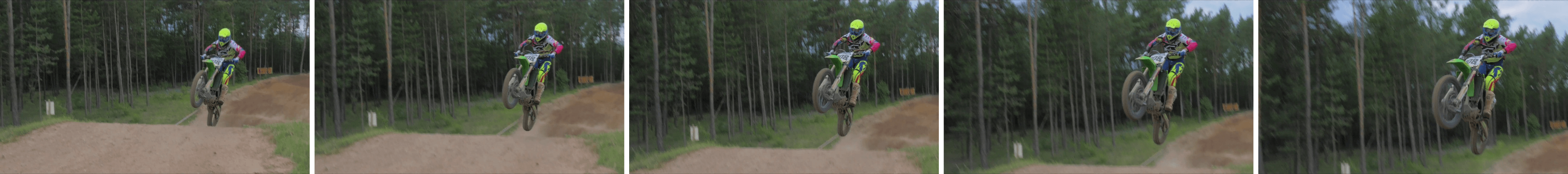}
        \caption{Unlabeled natural video sequence of DAVIS$_{17}$\cite{perazzi2016benchmark}val}
        \label{DAVIS_2017_picture}
    \end{subfigure}
    
    \begin{subfigure}[t]{\textwidth}
        \centering
        \includegraphics[width=\textwidth]{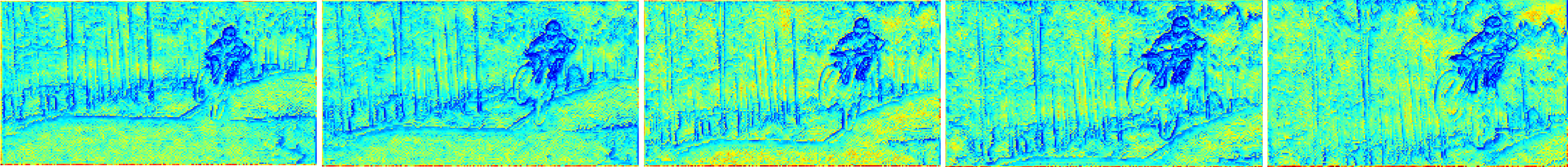}
        \caption{The feature maps obtained after extraction by the original encoder}
        \label{DAVIS_2017_originalfeature}
    \end{subfigure}
    
    \begin{subfigure}[t]{\textwidth}
        \centering
        \includegraphics[width=\textwidth]{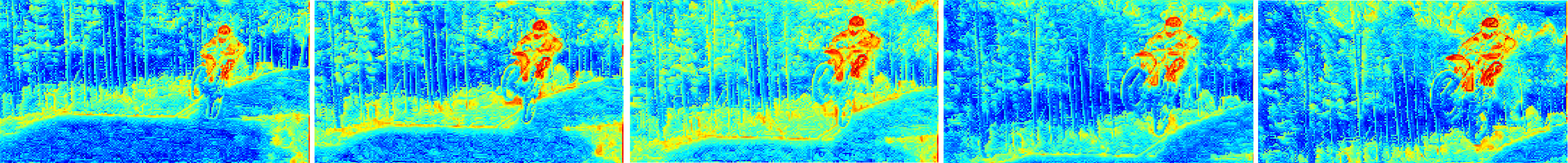}
        \caption{Motion-enhanced feature maps of this video sequence}
        \label{DAVIS_2017_ourfeature}
    \end{subfigure}
    \caption{The visualization results of our \textbf{Motion Enhancement Engine} framework.  These results strongly demonstrate that our algorithm can highlight the information of moving objects in frames (such as the motorcyclist, the foreground part within the forest area, etc.). See \ref{section 3.2 Motion Enhancement Engine} for details. }
    \label{visualization result of Motion Enhancement Engine}
\end{figure*}

\subsection{Motion Enhancement Engine} \label{section 3.2 Motion Enhancement Engine}
In the realm of self-supervised video learning, maintaining continuous focus on the target object can be a challenging endeavor, especially in the absence of labeled data.  Fortunately, videos inherently contain valuable motion information about the object, which can aid the feature encoder in consistently centering the camera on the target to the best of its ability. 

When dealing with the motion information of objects within a video, optical flow serves as a valuable tool.  A commendable approach involves channel fusion of optical flow maps from each frame, with an appropriate flow gap aligning with the current frame, before inputting them into the feature extractor for training (As in Fig.~\ref{fig:MER}(A) the part before the input Enc). 
This method effectively enhances the feature extractor's capacity to capture the characteristics of moving objects. The formula for channel fusion is given here:
\begin{equation}
    O_t = I_t + \alpha * F^{g}_{t}.
\end{equation}
Here, $I_t$ represents the frame $t$, $F$ signifies the optical flow map with a specific flow gap $g$ corresponding to the frame $t$, $\alpha$ is a learnable parameter, and $O$ denotes the result following channel fusion.  This fused output, is subsequently fed into the feature extractor $\Phi(O)$ for the next training step. 

While channel fusion undoubtedly empowers the feature extractor to capture increasingly valuable information, including intricate motion details that may be absent in individual frames, it also introduces certain inherent challenges.  For example, some moving background information may be incorrectly considered to high value.  Consequently, we introduce the \emph{Value Extraction Network} subsequent to the feature extraction (As shown in Fig.~\ref{fig:MER}(A) after Enc the part).  This mechanism guides the model during training to distinguish between moving targets and the background, allowing it to focus more on high-value areas.
\begin{figure}
    \centering
    \includegraphics[width=0.99\linewidth]{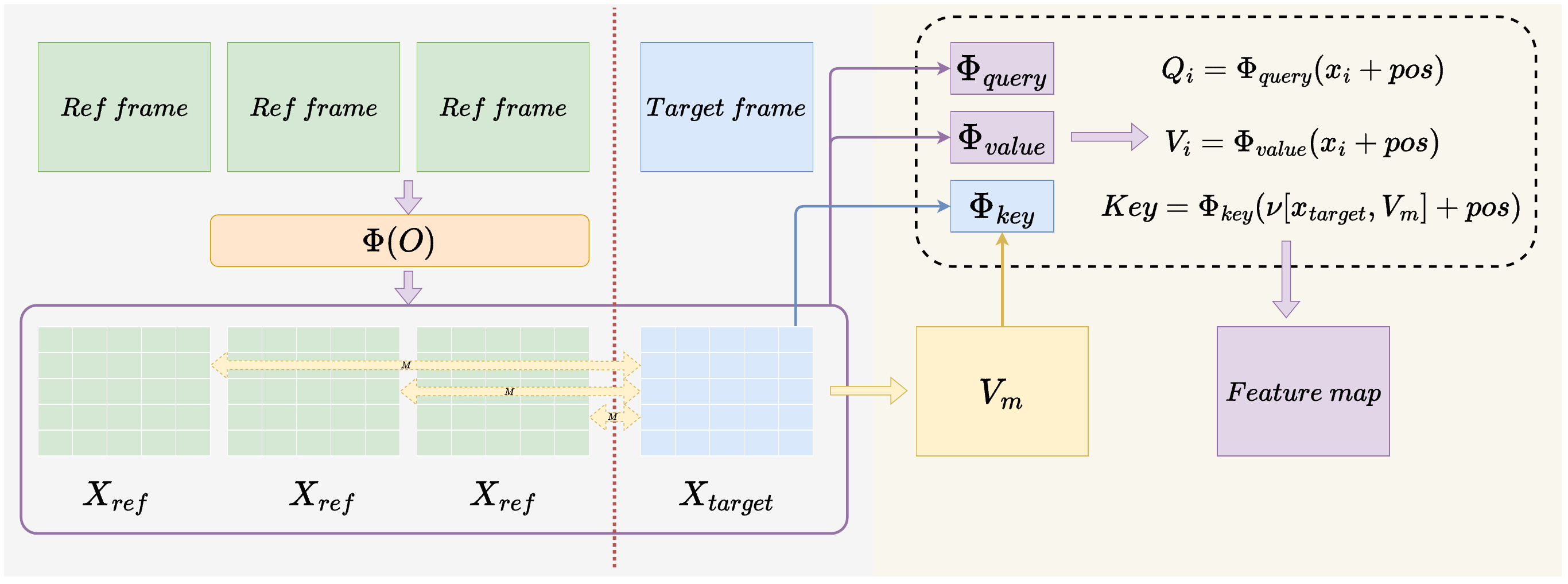}
    \caption{Our Value Extraction Network. Both $x_{target}$ and $x_{ref}$ are obtained through $\Phi(O)$.}
    \label{value fig}
\end{figure}

After obtaining the feature maps $x_{target}=\Phi(O_t)$ of target frame  and reference frames sequence $x_{ref}=[\Phi(O_{t-1}),\Phi(O_{t-2}),\cdot \cdot \cdot]$, we divide all frames into $N\times N$ smaller windows and match the similarity between the windows of $x_{target}$ and $x_{ref}$:
\begin{equation}
    [x_{target},x_{ref}] = [\Phi(O_{t}),\Phi(O_{t-1}),\Phi(O_{t-2}),\cdot \cdot \cdot],
\end{equation} 
\begin{equation}
    \begin{split}
    V^{i}_{m} = \sum_{ref}\sum_{j}^{N \times N} \textbf{M}(x^{i}_{t},x^{j}_{ref})-\Omega \textbf{M}(x^{i}_{t},x^{i}_{t}), 
    \end{split}
\end{equation}
Where,
\begin{equation}
    \textbf{M}(A,B)=\frac{{\sum A \times B}}{{\sqrt{\sum A^2} \times \sqrt{\sum B^2}}}.
\end{equation}
Here, $V_m\in \mathbb{R} ^{N\times N}$ represents the obtained potential value distribution map, $\Omega$ is a hyperparameter set manually within the range $[0.5, 1]$. The purpose of this step is to identify which regions in the target frame have appeared in the reference frames and consider them as potential regions where the tracked targets may be located. Next, we use $V_m$ to guide the model to focus more on high-value regions during the reconstruction process. Specifically, we embed $V_m$ into $x_{target}$ and then project it to a high-dimensional space through an encoder to obtain $Key$:
\begin{equation}
Key = \Phi_{key}(\nu [x_{target},V_m]+pos)\in \mathbb{R} ^{h\times w\times d}.
\end{equation}
Where, $\nu$ denotes the value map encoder, and $\Phi_{key}$ denotes a learnable encoder. In addition, to better distinguish between different regions, we added absolute positional information $pos\in \mathbb{R} ^{h\times w}$. Then, for each frame $x_i$ in $[x_{t},x_{t-1},x_{t-2},\cdot \cdot \cdot]$, we project it to the space $Q_i$ and $V_i$ with the same dimension as $Key$, and then operate with $Key$ to obtain the $Out$:
\begin{equation}
Q_i = \Phi_{query}(x_i+pos), V_i = \Phi_{value}(x_i+pos),
\end{equation}
\begin{equation}
    out = \text{softmax}\left(\frac{Q_i\times {Key}^T}{\sqrt{d_k}}\right)V_i,
\end{equation}
\begin{equation}
    Out = f(out) + x_i.
\end{equation}
Where, $Out$ denotes the result of the Value Extraction Network, and $f$ denotes the learnable function, $e. g$.  $f(\cdot) = \gamma * \cdot$.  During the training process, to prevent the model from focusing too much on a particular region and causing overfitting, we employ \emph{dropkey}\cite{li2023dropkey} to randomly mask a portion of $Key$. Fig.~\ref{value fig} illustrates this module. At the same time, we present the visualization results of feature extraction framework in Fig.~\ref{visualization result of Motion Enhancement Engine}. 

\subsection{Multi-Cluser Sampler} \label{section 3.3 Multi-Cluser Sampler}
In Section \ref{base}, we discussed the method for reconstructing pixels in the current frame by sampling from previous frames. The significance of selecting an appropriate similarity sampling method is crucial for achieving optimal results. To tackle this challenge, we propose a multi-scale, multi-iterative similarity sampling method. Specifically, each pixel in frame $I_{t}$ is compared to the pixels in frame $I_{ref},(ref \in [0,t-1])$ within a defined range centered on the corresponding location, ultimately generating a similarity matrix that captures the relationships between all pixels in frame $I_{t}$ and frame $I_{ref}$. 

However, the selection of the sampling range plays a crucial role in the similarity calculation. Considering that target pixels may resemble background pixels, an inappropriate choice of the sampling range for pixel similarity can lead to incorrect correlations between target and background pixels, resulting in inaccurate target tracking. This issue may manifest as a generated target mask that either over-covers the background or fails to adequately represent the target.

The above problems are often caused by inadequate or inaccurate extraction of similarity information. For example, a pixel representing a target in frame $I_{t}$ may be calculated to be very similar to a background pixel in frame $B$ because the sampling range is set too large, and thus be incorrectly assumed to be similar, when in fact the target pixel in frame $I_{ref}$ is more similar to the target pixel in frame $I_{t}$. On the contrary, if the sampling range is set too small, missing matches will be obvious, resulting in a loss of accuracy.

To address this problem, we propose an adaptive clustering sampling method based on region growth, termed Multi-Cluster Sampler. This method combines clustering techniques with region growth to achieve accurate sampling and tracking of target regions. 

First, we perform preliminary clustering on frames $I_{t}$ and $I_{ref}$ to distinguish the target from the background. Using the K-means clustering algorithm, we divide the pixels into clusters, generating initial outlines of the target and background regions. The cluster centers serve as starting points for region growth, laying the groundwork for subsequent similarity calculations. Based on clustering, we generate an initial similarity matrix $SimA_{init}$ through similarity sampling to characterize the similarity distribution of clusters in the current and previous frames. This initial sampling aims to obtain approximate information about the target region and create a base similarity frame. The construction formula for the initial similarity matrix can be expressed as:
\begin{equation}
    size_{origin} = \frac{1}{K} \sum_{i=1}^{K} \frac{1}{N_i} \sum_{j=1}^{N_i} d(c_i, p_j),
\end{equation}
\begin{equation}
    SimA_{init} = \Sigma(I_{t},I_{ref};size_{origin}).
    \label{SimA_init}
\end{equation}
Where $c_i$ is the clustering center of the $i$th cluster, $p_j$ is the $j$th point of the $i$th cluster, $d(c_i, p_j)$ denotes the distance between $c_i$ and $p_j$, and $size_{origin}$ is expressed as the average radius of multiple clusters obtained by K-means clustering, which serves as the initial sampling range (K serves as a hyperparameter and is set to be $5$ in this paper). From this the initial similarity matrix can be obtained after the sampler $\Sigma$.
\begin{figure}[ht]
    \centering
    \includegraphics[width=0.99\linewidth]{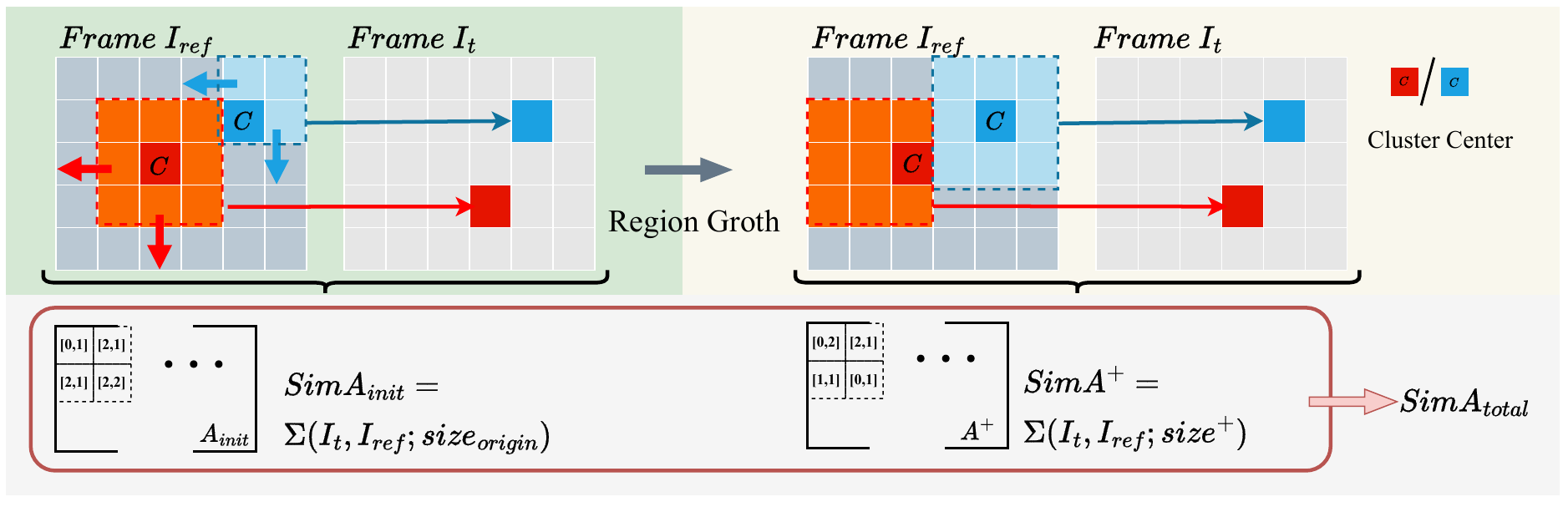}
    \caption{Illustration of the multi-scale similarity sampling method with adaptive region growth. Initial clustering distinguishes target and background regions, generating an initial similarity matrix \( SimA_{init} \). Region growth dynamically adjusts the sampling range based on similarity threshold \( \theta \), resulting in a refined similarity matrix \( SimA_{total} \) with enhanced target detail and background suppression.}
    \label{fig:multi_scale_sampling}
\end{figure}
Next, we perform a region growing operation on the initial similarity matrix. Region growing starts from the cluster center point of the initial clustering and gradually expands the sampling range according to the set similarity threshold $\theta$. Specifically, on the boundary of the current clustering, When the similarity between the boundary pixels and the clustering center is higher than the threshold $\theta$, the pixel is included in the sampling range; if the similarity is lower than $\theta$, the growth is stopped to avoid the introduction of background noise.

As the region growth proceeds, the sampling range will be dynamically adjusted according to the different distributions of the target regions. For regions with high similarity (i.e., target regions), the sampling range will be contracted to a smaller range to capture more detailed target features, while for background regions (i.e., regions with low similarity), the sampling range will be expanded appropriately to suppress the background details and emphasize the prominence of the target regions. In this way, the sampling range can be adapted to the variation of target and background regions in complex scenes.

After completing the region growth sampling, a similarity matrix $SimA_{total}$ containing richer information about the target region will be generated (See Fig. \ref{fig:multi_scale_sampling}), which is defined as follows:
\begin{equation}
    SimA_{total} = SimA_{init} + \delta SimA^{+} .
\end{equation}
Where $\Sigma$ is the similarity sampler, $SimA_{init}$ is the initial similarity matrix obtained from Eq. \ref{SimA_init}, and $SimA^{+} = \Sigma(I_{t},I_{ref};size^{+})$ is the sampling range after region growth.

\subsection{Fused loss function} \label{section 3.5 Fused loss function}
During the training process, our model comprises multiple sub-modules to handle different aspects of the object tracking task.  Since each of these parts has different evaluation criteria, we use a combination of the \emph{Positive-Negative samples reconstruction loss} and  \emph{Weight map loss} for loss calculation across different sub-modules.  

\subsubsection{Positive-Negative samples reconstruction loss} \label{section 3.5. 2 Positive-Negative samples reconstruction loss}
Previous research has indicated that when reconstructing a target frame from a reference frame, the required pixels typically come only from the current video sequence. Based on this characteristic, we have designed a contrastive learning mechanism to enhance the model's ability to select the correct pixels (As shown in Fig.~\ref{fig:MER}(C)). During the reconstruction process, pixels from the current video sequence and other video sequences are used as positive and negative samples, respectively, to reconstruct the current frame.  Concretely, given the query ${I}_{q}$ and negative samples reference frames ${I}_{n}(k^ {\prime})$ ,video affinity $A^{\prime}\in \left [ 0,1  \right ] ^{hw\times hw}$ is:
\begin{equation}
\begin{split}
    A^{\prime}(i, k)=\frac{\exp \left(\boldsymbol{I}_{q}(i) \cdot \boldsymbol{I}_{n}(k)\right)}{\sum_{n} \sum_{k^{\prime}} \exp \left(\boldsymbol{I}_{q}(i) \cdot \boldsymbol{I}_{n}(k^ {\prime})\right)}, i, k\in\left \{  1, \cdot \cdot \cdot , hw\right \} 
\end{split}
\end{equation}

In this process,we hope that if the model wrongly matches a query pixel $i$ with a negative but similar-looking pixel $k$ in $I_{n} $, the model can recognize that such pixel ${I}_{n}(k)$ is not the correct target to match with the current pixel ${I}_{q}(i)$, and consequently exclude it during the reconstruction process.  Therefore, we introduce the following loss function:
\begin{equation}
\mathcal{L}_{\mathrm{q}}=-\log \frac{\exp \left(q \cdot k_{+} / \tau\right)}{\sum_{i=0}^{k} \exp \left(q \cdot k_{i} / \tau\right)}
\end{equation}
Where $k_{+}$ represents the unique positive sample, and $k_{i}$ represents all the negative samples, $q$ represents the reconstructed frame we obtain. 
This loss function satisfies the following requirements: when $q$ is similar to the unique positive sample $k_{+}$ and dissimilar to all other negative samples $k_{i}$, the loss is low, and conversely, the loss is high.

\subsubsection{Weight map loss} \label{section 3.5. 3 Weight map loss}
In frame reconstruction processes, the smooth L1 loss is often employed to assess the loss between the reconstructed frame and the ground truth.  This loss function calculates the overall loss by computing the difference between each pixel $\hat{I}_{t}(i)$ in the reconstructed frame and the ground truth ${I}_{t}(i)$ and then taking the average. The loss function is expressed as follows:
\begin{equation}
\mathcal{L}=\frac{1}{n} \sum_{i} z_{i}
\end{equation}
where
\begin{equation}
z_{i}=\left\{\begin{array}{ll}
0. 5\left|\hat{I}_{t}(i)-{I}_{t}(i)\right|^{2}, & \text { if }\left|\hat{I}_{t}(i)-{I}_{t}(i)\right|<1 \\
\left|\hat{I}_{t}(i)-{I}_{t}(i)\right|-0. 5, & \text { otherwise }
\end{array}\right. 
\end{equation}

However, in the context of object tracking, the reconstruction quality of the target region in the frame should be the primary criterion for measuring the effectiveness of reconstruction. Therefore, we propose a specially tailored reconstruction loss function for object tracking. This loss function introduces a weight $\omega_{i}$ for each pixel, which assigns a higher weight to the loss in the target region during the computation, thus placing more emphasis on the reconstruction quality of the target area. The improved loss function is as follows:
\begin{equation}
\mathcal{L}_{\mathrm{m}}= \frac{1}{n} \sum_{i} z_{i} *\|\rho -\omega_{i} \|
\end{equation}

The overall weight map $\omega\in \left [ 0 ,2\right ] ^{h \times w}$ containing weights for each pixel is obtained by normalizing the result of  Value Extraction Network (See \ref{section 3.2 Motion Enhancement Engine} considered as background areas. 
The parameter $\rho$ is initialized as $0$ during the early stages of training, which biases the weight map $\omega$ towards the target, speeding up model convergence. As training progresses, $\rho$ gradually increases to $2$, causing $\omega$ focus more on the background, thus avoiding overfitting. 

Our overall loss consists of a combination of  Positive-Negative samples reconstruction loss $\mathcal{L}_{m}$ and Weight map loss $\mathcal{L}_{q}$ :
\begin{equation}
\mathcal{L}_{\mathrm{sum}}=  \mathcal{L}_\mathrm{m}+\gamma\mathcal{L}_\mathrm{q}
\end{equation}
Where $\gamma$ is a hyperparameter set by human intervention, determining the weighting of different components of the loss. 

\section{Experiment}\label{section 5 experiment}
We benchmark our model on DAVIS$_{17}$\cite{perazzi2016benchmark}val, in this benchmark, our model is required to propagate the first frame annotation to the entire video sequence. In \ref{section 5.2Diagnostic Experiment}, we conduct a set of ablative studies to examine the efficacy of our essential model design.

\subsection{Implementation Details}

\begin{figure}[h]
    \centering
    \begin{tikzpicture}
        \begin{axis}[
            width=0.9\columnwidth,
            height=4cm,
            every axis label/.append style={font=\small},
            every tick label/.append style={font=\footnotesize},
            xlabel={Epoch},
            ylabel={$\mathcal{J} \& \mathcal{F}$ (Mean)},
            xmin=1, xmax=20,
            ymin=64, ymax=80,
            xtick={1,5,10,15,20},
            ytick={65,67,69,71,73,75,77,79},
            grid=both,
            grid style={dotted},
            axis x line=bottom,
            axis y line=left,
            legend style={fill=none, draw=none, at={(1,1)}, anchor=north east}
        ]
            \addplot[smooth, mark=o, blue, semithick,
                mark options={fill=white, draw=blue, scale=1.0}]
            plot coordinates {
                (1,64.9)
                (2,68.0)
                (3,71.0)
                (4,72.2)
                (5,73.0)
                (6,73.4)
                (7,73.7)
                (8,73.9)
                (9,74.1)
                (10,74.3)
                (11,74.5)
                (12,74.7)
                (13,74.8)
                (14,75.0)
                (15,75.1)
                (16,75.2)
                (17,75.4)
                (18,75.8)
                (19,76.2)
                (20,76.4)
            };
            \addlegendentry{\scriptsize MER}
        \end{axis}
    \end{tikzpicture}
    \caption{Training performance of MER over 20 epochs on DAVIS$_{17}$\protect\cite{perazzi2016benchmark}val.}
    \label{fig:training_curve}
    \vspace{-0.5cm}
\end{figure}
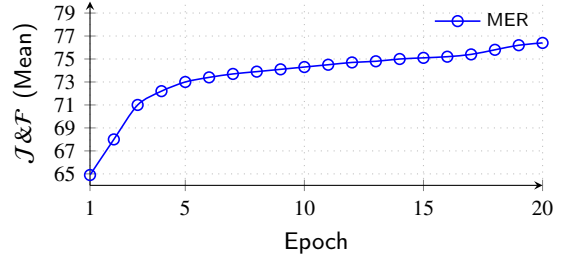

\noindent \textbf{Training:} To ensure a fair comparison, we used ResNet18 \cite{he2016deep} as the feature encoder in all experiments, generating feature embeddings with a spatial resolution of 1/4 the original image. The model underwent fully self-supervised training on two NVIDIA RTX3090s, starting with random weights and utilizing no external information beyond the original video sequence (YouTube-VOS\cite{xu2018youtube}). Frames were resized to $256\times256\times3$, applying Lab color space \cite{Lai_2020_CVPR} channel-level loss. Optimization used Adam with an initial learning rate of 0.001, dynamically adjusted based on iterations, and a batch size of 32. An online memory bank held 32$\times$50 frames from previous videos as negative samples. Metrics during training are shown in Fig.~\ref{fig:training_curve}

\begin{table*}[t]  %开始表格
\renewcommand\arraystretch{1.6} %调整行高
\centering
\begin{adjustbox}{width=\textwidth}
\begin{tabular}
{>{\centering\arraybackslash}m{3.0cm}
    >{\centering\arraybackslash}m{2.0cm}
    >{\centering\arraybackslash}m{3.0cm}
    >{\centering\arraybackslash}m{1.5cm}
    >{\centering\arraybackslash}m{1.5cm}
    >{\centering\arraybackslash}m{1.5cm}
    >{\centering\arraybackslash}m{1.5cm}
    >{\centering\arraybackslash}m{1.5cm}
    }
\toprule % 顶部横线
Method & Backbone & Dataset & $\mathcal{J}$\&$\mathcal{F}\uparrow$ & $\mathcal{J}$(Mean)$\uparrow$ & $\mathcal{J}$(Recall)$\uparrow$ & $\mathcal{F}$(Mean)$\uparrow$ & $\mathcal{F}$(Recall)$\uparrow$ \\

\midrule % 中间横线
MAST\cite{Lai_2020_CVPR}                  &ResNet-18   &Youtube-VOS          & 65.5 & 63.3 & 73.2 & 76.6 & 77.7 \\ % 表格内容
CRW\cite{jabri2020space}                  &ResNet-18   &Kinetics          & 68.3 &65.5 &78.6 &71.0 &82.9  \\
VFS\cite{xu2021rethinking}                &ResNet-18    &Kinetics             &66.7 &64.0 &- &69.4 &-\\
JSTG\cite{zhao2021modelling}              &ResNet-18    &Kinetics             &68.7 &65.8 &77.7 &71.6 &84.3\\
CLTC\cite{jeon2021mining}                 &ResNet-18    &Youtube-VOS        &70.3 &67.9 &78.2 &72.6 &83.7 \\
DINO\cite{caron2021emerging}              &ViT-B/8      &ImageNet            &71.4 &67.9 &- &74.9& -\\
LIIR\cite{li2022locality}                 &ResNet-18    &Youtube-VOS       &72.1 &69.7 &81.4 &74.5 &85.9\\
MASk-VOS\cite{li2023unified}              &ResNet-18    &Youtube-VOS        &74.5 &71.6   &82.9   &77.4   &86.9\\
\hline
\textbf{MER}                              &ResNet-18    &Youtube-VOS         &\textbf{76.5} &\textbf{74.8} &\textbf{87.6} &\textbf{78.2} &\textbf{89.1}\\
\bottomrule % 底部横线
\end{tabular}
\end{adjustbox}
\caption{\textbf{Quantitative results for video object segmentation} on DAVIS$_{17}$\protect\cite{perazzi2016benchmark}val.}
\label{Quantitative results for DAVIS}
\end{table*}  %结束表格

\noindent \textbf{Testing:} Following the training phase, the model is directly applied to downstream tasks. During testing, we integrate long-term memory (e.g., $I_0$ and $I_5$) with short-term memory (e.g., $I_{t-5}$, $I_{t-3}$, and $I_{t-1}$) to establish correspondences between the query frame and key frames. Our experiments demonstrate that the selection of frames has a minimal impact on performance; however, the concurrent utilization of both long-term and short-term memories proves to be essential.

\begin{figure*}[h]
    \centering
    \begin{subfigure}[t]{\textwidth}
        \centering
        \includegraphics[width=\textwidth]{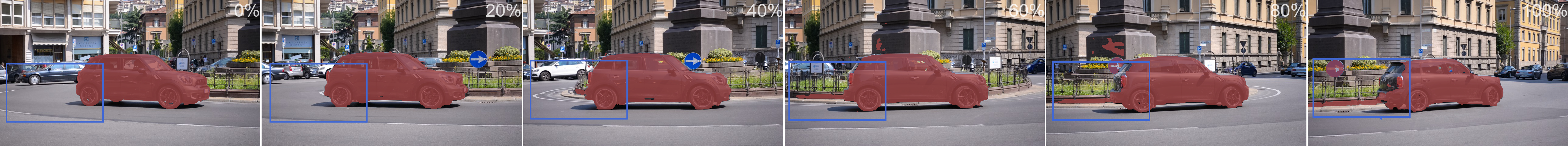}
        \caption{Results of video object segmentation task conducted without introducing our framework}
        \label{car_LIIR}
    \end{subfigure}    
    
    \begin{subfigure}[t]{\textwidth}
        \centering
        \includegraphics[width=\textwidth]{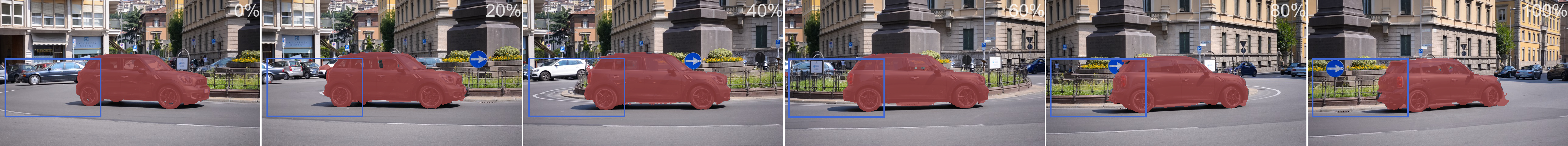}
        \caption{Results of the video object segmentation task conducted by MER}
        \label{car_our}
    \end{subfigure}
    \caption{The difference in results for video object segmentation (Section \ref{section 5 experiment}) on DAVIS$_{17}$\protect\cite{perazzi2016benchmark}val between without our framework(top) and \textbf{MER}(bottom). The interval between each image is 15 frames.}
    \label{visible benchmark}
\end{figure*}

\begin{figure*}[h]
    \centering
    \includegraphics[width=0.99\linewidth]{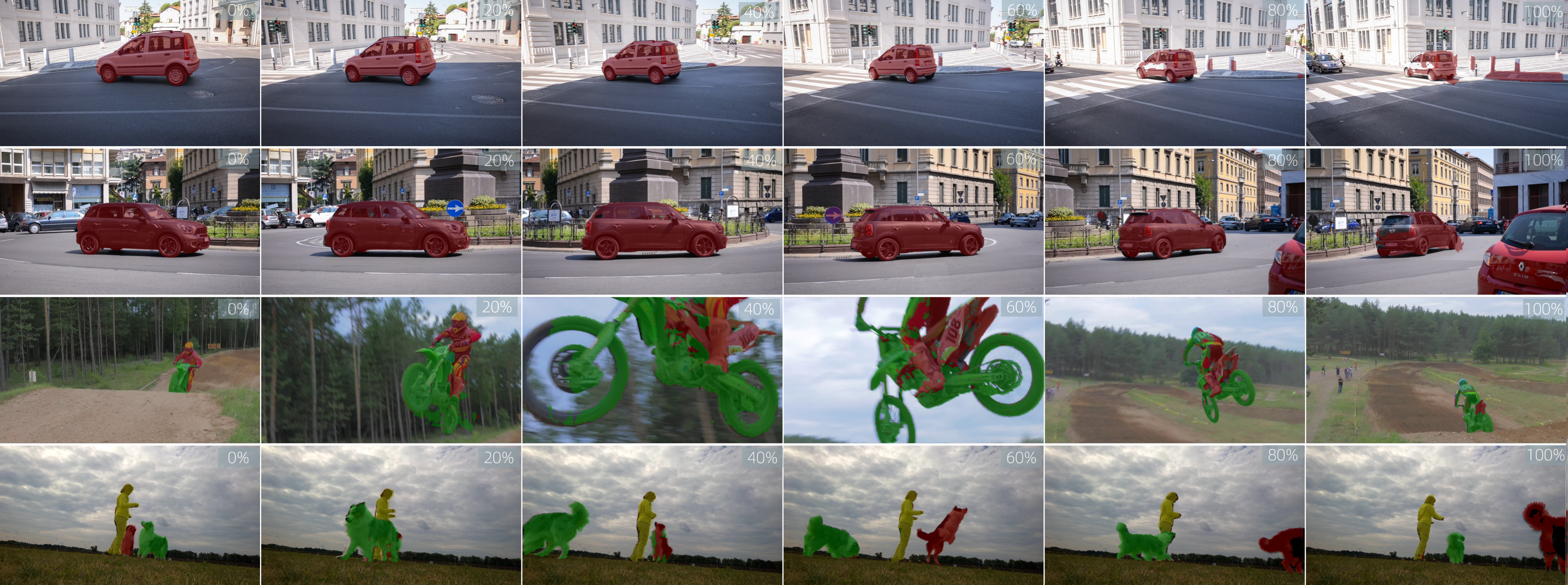}
    \caption{Visualization of the segmentation effect of the \textbf{MER} on DAVIS$_{17}$\protect\cite{perazzi2016benchmark}val}
    \label{fig:visualize}
\end{figure*}

\subsection{Video Segmentation on DAVIS-2017} \label{section 5.1 Video Segmentation on DAVIS-2017}
In Table \ref{Quantitative results for DAVIS}, we compare our model with previous approaches on the DAVIS$_{17}$\cite{perazzi2016benchmark}val. We assess methods for semi-supervised video object segmentation using three evaluation metrics: the mean region similarity $\mathcal{J}(Mean)$, the mean contour accuracy $\mathcal{F}(Mean)$, and their average \( J \& F \).

It can be observed that our model outperforms all current self-supervised object tracking methods in the benchmark test. It surpasses current best-performing self-supervised method, $i.e.$, MASK-VOS\cite{li2023unified} in terms of mean $\mathcal{J}$\&$\mathcal{F}$ (\textbf{76.5} vs. 74.5). Furthermore, our model exhibits a stronger capability to distinguish between the target and the background, resulting in improved tracking performance. Fig.~\ref{visible benchmark} supports our conclusion, compared to other target tracking models (Fig.~\ref{visible benchmark}(A)), MER (Fig.~\ref{visible benchmark}(B)) prevents target annotation propagates to the background ($e.g.$, the rear part of the car in the image). See Fig.~\ref{fig:visualize} for more visualizations.

\begin{table}[h]  %开始表格
\renewcommand\arraystretch{1.4} %调整行高
\centering
\begin{adjustbox}{width=0.99\columnwidth}
\begin{tabular}
{>{\centering\arraybackslash}m{3cm}
    >{\centering\arraybackslash}m{2cm}
    >{\centering\arraybackslash}m{3cm}
    >{\centering\arraybackslash}m{2cm}
    }
\toprule % 顶部横线
Method &Backbone & Dataset & $\mathcal{J}$\&$\mathcal{F}(Mean)\uparrow$  \\
\midrule % 中间横线
MAST\cite{Lai_2020_CVPR}  &ResNet-18    &Youtube-VOS &64.2 \\
CLTC\cite{jeon2021mining}  &ResNet-18    &Youtube-VOS &67.3 \\
LIIR\cite{li2022locality} &ResNet-18    &Youtube-VOS  &69.3 \\
MASk-VOS\cite{li2023unified} &ResNet-18  & - &71.6 \\
\hline
\textbf{MER}  &ResNet-18     &Youtube-VOS         &\textbf{73.5} \\
\bottomrule % 底部横线
\end{tabular}
\end{adjustbox}
\caption{\textbf{Quantitative results for video object segmentation} on YouTube-VOS\cite{xu2018youtube}val. }
\label{Quantitative results for ytb}
\end{table}  %结束表格

\subsection{Video Segmentation on YouTube-VOS} \label{section 5.1 Video Segmentation on YouTube-VOS}
In Table \ref{Quantitative results for ytb}, we compare our model with previous approaches on the YouTube-VOS\cite{xu2018youtube}val. We also assess methods for semi-supervised video object segmentation using three evaluation metrics: the mean region similarity $\mathcal{J}(Mean)$, the mean contour accuracy $\mathcal{F}(Mean)$, and their average \( J \& F (Mean)\).

It can be observed that our model outperforms all current self-supervised object tracking methods in the benchmark test. It surpasses current best-performing self-supervised method, $i.e.$, MASK-VOS\cite{li2023unified} in terms of mean $\mathcal{J}$\&$\mathcal{F}$ (\textbf{73.5} vs. 71.6). Furthermore, our model exhibits a stronger capability to distinguish between the target and the background, resulting in improved tracking performance. 

\begin{figure*}[h]
    \centering
    \includegraphics[width=0.99\textwidth]{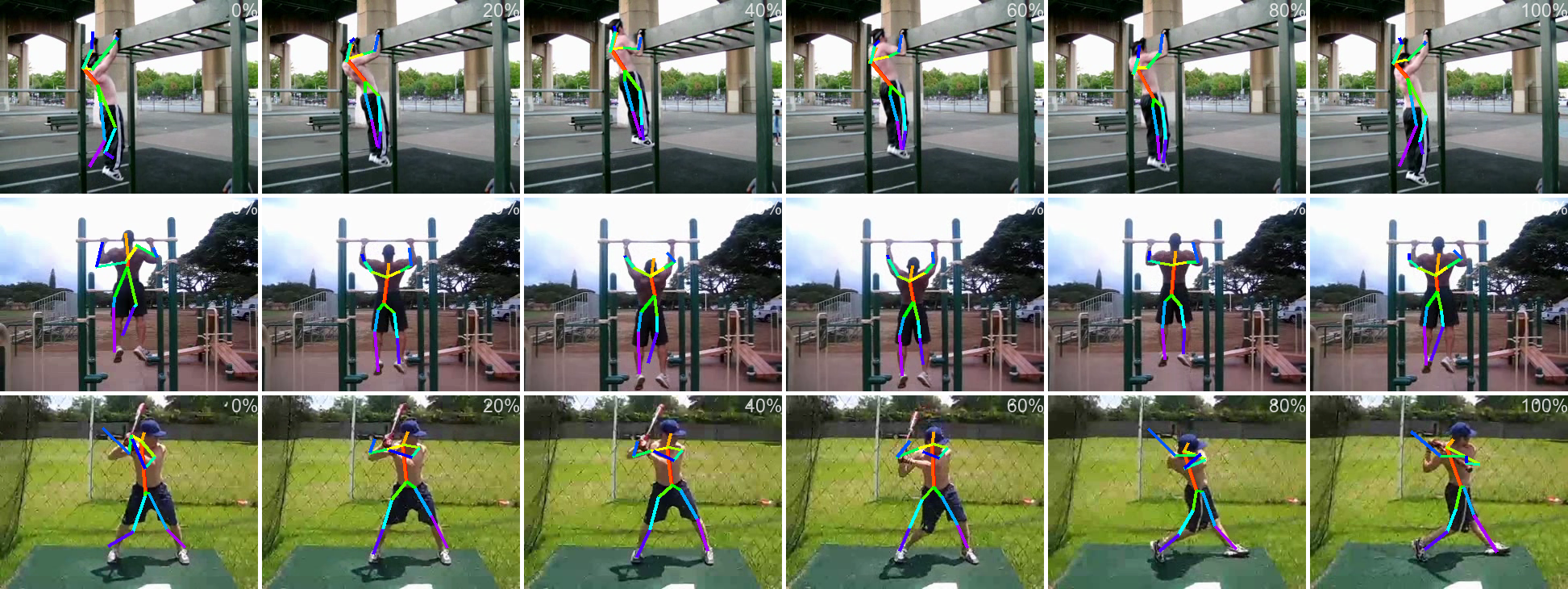}
    \caption{\textbf{Visualization results for keypoint propagation} on JHMDB\protect\cite{jhuang2013towards}val}
    \label{jhmdb_png}
\end{figure*}

\begin{table}[h]  %开始表格
\renewcommand\arraystretch{1.4} %调整行高
\centering
\begin{adjustbox}{width=0.99\columnwidth}
\begin{tabular}
{>{\centering\arraybackslash}m{3cm}
    >{\centering\arraybackslash}m{1.2cm}
    >{\centering\arraybackslash}m{4cm}
    >{\centering\arraybackslash}m{2cm}
    }
\toprule % 顶部横线
Method &Supervised & Dataset & $PCK@0.1\uparrow$  \\

\midrule % 中间横线
CRW\cite{jabri2020space}  &    &Kinetics &59.3 \\
VFS\cite{xu2021rethinking} &   &Kinetics  &60.5 \\
JSTG\cite{zhao2021modelling}  &    &Kinetics  &61.4 \\
CLTC\cite{jeon2021mining} &    &Youtube-VOS  &60.5 \\
LIIR\cite{li2022locality} &    &Youtube-VOS  &60.7 \\
SPAT\cite{li2023spatial}  &    &ImageNet+Youtube-VOS &63.1 \\
\hline
\textbf{MER}  &     &Youtube-VOS          &\textbf{64.3} \\
\hline
ResNet-18\cite{he2016deep}  &$\surd$   &- &53.8 \\
Thin-Slicing Net\cite{song2017thin} &$\surd$  &- &68.7 \\
\bottomrule % 底部横线
\end{tabular}
\end{adjustbox}
\caption{\textbf{Quantitative results for human part propagation and
pose keypoint tracking}. We show results of self-supervised methods and some supervised methods for comparison.}
\label{Quantitative results for JHMDB}
\end{table}  %结束表格

\subsection{Pose Keypoint Tracking on JHMDB}
Table \ref{Quantitative results for JHMDB} presents the outcomes obtained by our model (MER) on the JHMDB dataset \cite{jhuang2013towards}, showcasing the quantitative results of human part propagation and pose keypoint tracking. We employ the standard PCK\cite{yang2012articulated} for all visible points, rather than computing the PCK for each video individually and then averaging the results, as our evaluation metric. Several tracking visualization outcomes are depicted in Fig.~\ref{jhmdb_png}. Notably, our model surpasses the majority of prior self-supervised object tracking models, achieving state-of-the-art (SOTA) results particularly on the $PCK@0.1$ metric. Moreover, our work demonstrates significant enhancements in performance compared to certain partially supervised object tracking models.

\subsection{Hyperparameter Experiments}
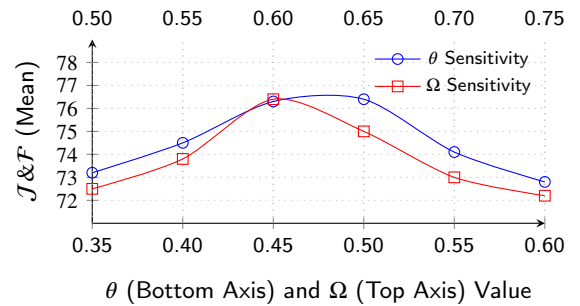
\begin{figure}[h]
    \centering
    \begin{tikzpicture}
        \begin{axis}[
            width=0.9\columnwidth,
            height=4cm,
            every axis label/.append style={font=\small},
            every tick label/.append style={font=\footnotesize},
            xlabel={$\theta$ (Bottom Axis) and $\Omega$ (Top Axis) Value },
            ylabel={$\mathcal{J} \& \mathcal{F}$ (Mean)},
            xmin=0, xmax=1,
            ymin=71, ymax=79,
            xtick={0,0.2,0.4,0.6,0.8,1.0},
            xticklabels={0.35,0.40,0.45,0.50,0.55,0.60}, % Evenly spaced labels for beta
            ytick={72,73,74,75,76,77,78},
            legend pos=south east,
            grid=both,
            grid style={dotted},
            axis y line=left,
            axis x line=bottom,
            extra x ticks={0,0.2,0.4,0.6,0.8,1.0}, % Evenly spaced extra x ticks for gamma
            extra x tick labels={0.50,0.55,0.60,0.65,0.70,0.75}, % Labels for gamma
            extra x tick style={ticklabel pos=top},
            legend style={fill=none, draw=none, at={(1,1)}, anchor=north east}
        ]
        
        % Beta sensitivity curve
        \addplot[smooth,mark=o,blue] plot coordinates {
            (0, 73.2)
            (0.2, 74.5)
            (0.4, 76.3)
            (0.6, 76.4)
            (0.8, 74.1)
            (1, 72.8)
        };
        \addlegendentry{\scriptsize\textcolor{black}{$\theta$ Sensitivity}}

        % Gamma sensitivity curve
        \addplot[smooth,mark=square,red] plot coordinates {
            (0, 72.5)
            (0.2, 73.8)
            (0.4, 76.4)
            (0.6, 75.0)
            (0.8, 73.0)
            (1, 72.2)
        };
        \addlegendentry{\scriptsize\textcolor{black}{$\Omega$ Sensitivity}}

        \end{axis}
    \end{tikzpicture}
    \caption{Impact of hyperparameters $\theta$ and $\Omega$ on $\mathcal{J} \& \mathcal{F}$ (Mean). The bottom x-axis represents $\theta$, and the top x-axis represents $\Omega$, both evenly spaced.}
    \label{fig:beta_gamma_sensitivity}
    \vspace{-0.5cm}
\end{figure}

We treat the similarity threshold $\theta$ of the Multi-Cluster Sampler and the hyperparameter $\Omega$ of the Value Extraction Network as adaptive parameters. The threshold $\theta$ is initially determined based on an in-depth analysis of the dataset with a specific focus on maintaining temporal consistency across video frames, which is crucial for grouping similar features accurately. In contrast, $\Omega$ regulates the degree of feature aggregation during training, ensuring robust and stable pixel matching across frames. Extensive initialization tuning experiments, as illustrated in Figure~\ref{fig:beta_gamma_sensitivity}, validate that setting $\theta$ to 0.45 and $\Omega$ to 0.6 achieves an optimal balance, leading to the best overall performance.

\subsection{Ablation Experiment} \label{section 5.2Diagnostic Experiment}
To investigate the impact of different modules, we conducted a series of ablative studies by systematically removing one module at a time. In these experiments, all models were trained on YouTube-VOS\cite{xu2018youtube}, and evaluated on DAVIS$_{17}$\cite{perazzi2016benchmark}val.

\begin{table}[h]
\renewcommand\arraystretch{1.2}
\centering  
\begin{adjustbox}{width=\columnwidth}
\begin{tabular}
    {>{\centering\arraybackslash}m{3cm}
    >{\centering\arraybackslash}m{1.6cm}
    >{\centering\arraybackslash}m{1.6cm}
    >{\centering\arraybackslash}m{0.8cm}
    >{\centering\arraybackslash}m{0.8cm}
    >{\centering\arraybackslash}m{1cm}}
\toprule
            \multicolumn{1}{c}{Motion-enhanced }        & \multicolumn{1}{c}{Multi-Cluser}     & \multicolumn{1}{c}{Fused}& \multicolumn{3}{c}{DAVIS} \\
\cline{4-6} \multicolumn{1}{c}{feature extraction}      & \multicolumn{1}{c}{Sampler}  & \multicolumn{1}{c}{loss function}&
            \multicolumn{1}{c}{ $\mathcal{J}\uparrow$} & \multicolumn{1}{c}{$\mathcal{F}\uparrow$}  &\multicolumn{1}{c}{$\mathcal{J} \&\mathcal{F}\uparrow$}\\
            \hline
            \hline
               -        &   -       &-         & 68.8     & 72.6  &70.7\\
            \hline
            $\surd$     &   -       &-         & 71.7     & 75.2  &73.4\\
               -        & $\surd$   &-         & 72.3     & 74.7  &73.5\\
               -        & -         &$\surd$   & 71.2     & 73.1  &72.2\\
            $\surd$     & $\surd$   &$\surd$   & 74.8     & 78.2  &76.5\\
            \bottomrule
\end{tabular}
\end{adjustbox}
\caption{Detailed analysis of essential components of MER on DAVIS$_{17}$\protect\cite{perazzi2016benchmark}val. See \ref{section 5.2.1 Key Component Analysis} for details.} 
\label{All detail}
\end{table}

\subsubsection{Key Component Analysis} \label{section 5.2.1 Key Component Analysis}
We first examine the efficacy of essential components of MER, $i.e.$, Motion Enhancement Engine, Multi-Cluser Sampler, and Fused loss function. The results are summarized in Table \ref{All detail}. In lines 2-4, we demonstrate the individual impact of using specific modules on the overall model performance. In line 5, we combine all parts to create the MER and achieve the best performance. This indicates that these modules can complement each other and work synergistically, showcasing the effectiveness of our overall design.

\begin{table}[h]
\renewcommand\arraystretch{1.2}
    \begin{subtable}{0.49\textwidth}
        \centering
        \begin{adjustbox}{width=\textwidth}
        \begin{tabular}
        {>{\centering\arraybackslash}m{3cm}
        >{\centering\arraybackslash}m{3cm}
        >{\centering\arraybackslash}m{1.5cm}
        >{\centering\arraybackslash}m{1.5cm}}
            \toprule
                            \multicolumn{1}{c}{Optical} & \multicolumn{1}{c}{Value}  & \multicolumn{2}{c}{DAVIS} \\
            \cline{3-4}     \multicolumn{1}{c}{flow}  & \multicolumn{1}{c}{extraction network}  & \multicolumn{1}{c}{ $\mathcal{J}\uparrow$} & \multicolumn{1}{c}{$\mathcal{F}\uparrow$} \\
            \hline
            \hline
                -       &   -         & 69.8     & 74.0 \\
            \hline
            $\surd$     &  -          & 72.1     & 75.7 \\
               -        & $\surd$     & 72.9    & 77.1 \\
            $\surd$     & $\surd$     & 74.8     & 78.2 \\
            \bottomrule
        \end{tabular}
        \end{adjustbox}
        \caption{Motion Enhancement Engine}
        \label{Motion Enhancement Engine}
    \end{subtable}
    \begin{subtable}{0.49\textwidth}
        \centering
        \begin{adjustbox}{width=\textwidth}
        \begin{tabular}
        {>{\centering\arraybackslash}m{3cm}
        >{\centering\arraybackslash}m{3cm}
        >{\centering\arraybackslash}m{1.5cm}
        >{\centering\arraybackslash}m{1.5cm}}
            \toprule
                            \multicolumn{1}{c}{Num of Cluster} & \multicolumn{1}{c}{Multi-Cluser}  & \multicolumn{2}{c}{DAVIS} \\
            \cline{3-4}     \multicolumn{1}{c}{Centers}    & \multicolumn{1}{c}{Sampler}  & \multicolumn{1}{c}{ $\mathcal{J}\uparrow$} & \multicolumn{1}{c}{$\mathcal{F}\uparrow$} \\
            \hline
            \hline
            -     &   -         & 70.7     & 74.4 \\
            \hline
            3     & $\surd$     & 72.6     & 76.1 \\
            7     & $\surd$     & 73.0     & 76.7 \\
            5     & $\surd$     & 74.8     & 78.2 \\
            \bottomrule            
        \end{tabular}
        \end{adjustbox}
        \caption{Multi-Cluser Sampler}
        \label{Multi-Cluser Sampler}
    \end{subtable}
    \begin{subtable}{0.49\textwidth}
        \centering
        \begin{adjustbox}{width=\textwidth}
        \begin{tabular}
        {>{\centering\arraybackslash}m{3cm}
        >{\centering\arraybackslash}m{3cm}
        >{\centering\arraybackslash}m{1.5cm}
        >{\centering\arraybackslash}m{1.5cm}}
            \toprule
                            \multicolumn{1}{c}{Weight map} & \multicolumn{1}{c}{Negative-Positive}  & \multicolumn{2}{c}{DAVIS} \\
            \cline{3-4}     \multicolumn{1}{c}{loss}    & \multicolumn{1}{c}{samples loss}  & \multicolumn{1}{c}{ $\mathcal{J}\uparrow$} & \multicolumn{1}{c}{$\mathcal{F}\uparrow$} \\
            \hline
            \hline
               -        &   -         & 70.8     & 74.9\\
            \hline
            $\surd$     &   -         & 73.1     & 74.9 \\
               -        & $\surd$     & 73.5     & 75.2 \\
            $\surd$     & $\surd$     & 74.8     & 78.2 \\
            \bottomrule
        \end{tabular}
        \end{adjustbox}
        \caption{Fused loss function}
        \label{Fused loss function}
    \end{subtable}
    \caption{A set of ablative studies on DAVIS$_{17}$\protect\cite{perazzi2016benchmark}val. See \ref{section 5.2Diagnostic Experiment} for details.} 
    \label{ablative studies}
\end{table}

\subsubsection{Motion Enhancement Engine} \label{section 5.2.2 Motion Enhancement Engine}
We explored two approaches to emphasize the most valuable regions of the video, specifically moving targets. First, before the image enters the feature extractor, we enhance target visibility by superimposing the optical flow map onto the image, effectively distinguishing between static and dynamic objects. Second, after obtaining the feature map, we introduce a value extraction network to guide the model’s attention toward high-value moving targets. As shown in Table \ref{ablative studies}(A), both modules function effectively on their own. \textbf{Notably}, when the value extraction network operates without the optical flow-enhanced feature map, it behaves similarly to a self-attention mechanism, still improving the model’s ability to extract target features and enhancing segmentation accuracy. However, the network achieves optimal performance only when it receives the optical flow-fused feature map, demonstrating the synergy between the two modules.

\subsubsection{Multi-Cluser Sampler} \label{section 5.2.3 Multi-Cluser Sampler}
Table \ref{ablative studies}(B) shows the performance of the model under different conditions, and it can be seen that the model can still improve the segmentation effect without category labels (we have no way of knowing the class of the actual segmentation target). The selection of category hyperparameters should not be too small or too large, too small will cause the sampler's matching receptive field to be very small, and too large will increase the probability of false matching, which is consistent with the assumption.

\subsubsection{Fused loss function} \label{section 5.2.4 Fused loss function}
As a baseline, the model uses the smooth L1 loss and spatial compactness prior loss. Building upon this, we consider three different combinations of loss functions during the training process to evaluate the effectiveness of Fused loss function (Section \ref{section 3.5 Fused loss function}):$(i)$ Adding Positive and Negative samples reconstruction loss, without Weight map loss. $(ii)$ Replace smooth L1 loss with the weight map loss, without introducing Positive and Negative samples reconstruction loss. $(iii)$ Using both of them. As shown in Table \ref{ablative studies}(C), the loss functions we designed enable the model to perform more accurate result evaluations.

\section{Conclusion}
In summary, our proposed MER method provides a concise yet effective enhancement for self-supervised Video Correspondence Learning through three key contributions. \textbf{First}, MER amplifies the motion details of the target within the video, employing the motion as an anchoring reference to meticulously extract pertinent, high-value information from the video frames. This strategy enables a more precise and efficient capture of the target's pixel features. \textbf{Secondly}, during the pixel sampling reconstruction phase, we employ Multi-Cluser Sampler technology to rectify inaccuracies in pixel correspondences across video frames, while simultaneously enriching the process with additional information conducive to pixel reconstruction. \textbf{Finally}, by integrating a video frame storage library, employing positive and negative sample comparison loss, and utilizing a weight graph constraint model, our model enriches its analytical framework with substantial information, all while consistently maintaining precise focus on the target. \textbf{Notably}, trained without semantic annotations, our model outperforms prior self-supervised methods on both DAVIS$_{17}$\protect\cite{perazzi2016benchmark}val (\emph{Increased by 2.7\%}), a benchmark dataset for video object segmentation, and JHMDB\protect\cite{jhuang2013towards}val (\emph{Increased by 1.9\%}), a benchmark dataset for pose keypoint tracking. This advancement highlights the great potential for enhancing video analytics without relying on semantic annotations.

\section*{Acknowledgments}
This research was supported in part by National Natural Science Foundation of China under grant No. 62061146001, 62372102, 62232004, 61972083, 62202096, 62072103, Jiangsu Provincial Key Research and Development Program under grant No. BE2022680, Fundamental Research Funds for the Central Universities (242024k30022), Jiangsu Provincial Key Laboratory of Network and Information Security under grant No. BM2003201, and Collaborative Innovation Center of Novel Software Technology and Industrialization.

\printcredits

\bibliographystyle{cas-model2-names}

\bibliography{cas-dc-template}

\vskip 3pt

\bio{authors/zihanzhou}
\textbf{Zihan Zhou} received the B.S. degree from Shanghai Dianji University, and is currently pursuing the M.S. degree in the School of Computer Science and Engineering at Southeast University, China, since 2022. His research interests include self-supervised video processing and multimodal video generation within the field of computer vision.
\endbio
\vspace{1cm}

\bio{authors/changruidai}
\textbf{Changrui Dai} received the B.S. degree from Nanjing Agricultural University, and is currently pursuing the M.S. degree in the School of Computer Science and Engineering at Southeast University, China, since 2022. His research interests include self-supervised video processing and deep neural network design.
\endbio
\vspace{1cm}

\bio{authors/aibosong}
\textbf{Dr. Aibo Song} received the M.S. degree from Shandong University of Science and Technology, and the Ph.D.degree from Southeast University, China, in 1996 and 2003 respectively. He is currently a professor in the School of Computer Science and Engineering, Southeast University. His current research interests include big data processing, and cloud computing.
\endbio
\vspace{1cm}

\bio{authors/xiaolinfang}
\textbf{Dr. Xiaolin Fang} received the B.S. degree from Harbin Engineering University, Harbin, China, in 2007, and the M.S. and Ph.D. degrees from the Harbin Institute of Technology, Harbin, China, in 2009 and 2014 respectively. He is currently an Associate Professor with the School of Computer Science and Engineering, Southeast University, Nanjing, China. His research interests include sensor networks, data processing, image processing, and scheduling.
\endbio

\end{document}